\documentclass[letterpaper]{article} 
\usepackage{aaai24}  
\usepackage{times}  
\usepackage{helvet}  
\usepackage{courier}  
\usepackage[hyphens]{url}  
\usepackage{graphicx} 
\usepackage{natbib}  
\usepackage{caption} 
\usepackage{algorithm}
\usepackage{algorithmic}

\usepackage{amssymb}
\usepackage{amsmath}
\usepackage{newfloat}
\usepackage{listings}
\DeclareCaptionStyle{ruled}{labelfont=normalfont,labelsep=colon,strut=off} 
\floatstyle{ruled}
\newfloat{listing}{tb}{lst}{}
\floatname{listing}{Listing}
\usepackage{multirow}

\title{Enhancing Low-Resource Relation Representations through Multi-View Decoupling}
\author{
    Chenghao Fan\textsuperscript{\rm 1,\rm 2},
    Wei Wei\thanks{*Corresponding author}\textsuperscript{\rm 1,\rm 2},
    Xiaoye Qu\textsuperscript{\rm 1,\rm 2,\rm 5},
    Zhenyi Lu\textsuperscript{\rm 1,\rm 2}, \\
    Wenfeng Xie\textsuperscript{\rm 3},
    Yu Cheng\textsuperscript{\rm 4},
    Dangyang Chen\textsuperscript{\rm 3}\\
}
\affiliations{
 \textsuperscript{\rm 1}Cognitive Computing and Intelligent Information Processing (CCIIP) Laboratory, School of Computer Science and Technology, Huazhong University of Science and Technology \\
\textsuperscript{\rm 2}Joint Laboratory of HUST and Pingan Property \& Casualty Research (HPL) \\
\textsuperscript{\rm 3}Ping An Property \& Casualty Insurance Company of China, Ltd. \\
\textsuperscript{\rm 4}The Chinese University of Hong Kong\\
\textsuperscript{\rm 5}Shanghai AI Laboratory\\
facicofan@gmail.com,\{weiw,quxiaoye\}@hust.edu.cn,luzhenyi529@gmail.com,xiewenfeng801@pingan.com.cn\\chengyu@cse.cuhk.edu.hk,chendangyang273@pingan.com.cn
}

\usepackage{bibentry}
\usepackage{xcolor}
\usepackage{hyperref}

\hypersetup{
    colorlinks=true,
    linkcolor=blue,
    filecolor=magenta,      
    urlcolor=cyan,
    citecolor=blue,
}
\begin{document}

\maketitle

\begin{abstract}
Recently, prompt-tuning with pre-trained language models (PLMs) has demonstrated the significantly enhancing ability of relation extraction (RE) tasks. 
However, in low-resource scenarios, where the available training data is scarce, previous prompt-based methods may still perform poorly for prompt-based representation learning due to a superficial understanding of the relation. 
To this end, we highlight the importance of learning high-quality relation representation in low-resource scenarios for RE, and propose a novel prompt-based relation representation method, named MVRE (\underline{M}ulti-\underline{V}iew \underline{R}elation \underline{E}xtraction), to better leverage the capacity of PLMs to improve the performance of RE within the low-resource prompt-tuning paradigm. Specifically, MVRE decouples each relation into different perspectives to encompass multi-view relation representations for maximizing the likelihood during relation inference.
Furthermore, we also design a Global-Local loss and a Dynamic-Initialization method for better alignment of the multi-view relation-representing virtual words, containing the semantics of relation labels during the optimization learning process and initialization. Extensive experiments on
three benchmark datasets show that our method can achieve
state-of-the-art in low-resource settings. The code is available at \url{https://github.com/Facico/MVRE}.
\end{abstract}

\section{Introduction}

Relation Extraction (RE) aims to extract the relation between two entities \cite{qu2023distantly,gu2022delving} from an unstructured text \cite{cheng2021hacred}. Given the significance of inter-entity relations within textual information, the practice of relation extraction finds extensive utility across various downstream tasks, including dialogue systems~\cite{lu2023miracle,liu2018knowledge}, information retrieval~\cite{yang2020biomedical,yu2023fusionint5}, information extraction \cite{zhu2023mirror,zhu2021efficient}, and question answering~\cite{yasunaga2021qa,qu2021passage}.

Following the emergence of the paradigm involving pre-trained models and fine-tuning for downstream tasks~\cite{kenton2019bert,radford2018improving}, many recent relation extraction studies have embraced the utilization of large language  models~\cite{ye2020coreferential,soares2019matching,zhou2022improved,ye2022packed}. In these works, the language models are integrated with classification heads and fine-tuned specifically for relation extraction tasks, resulting in promising results.
However, the effective training of additional classification heads becomes challenging in situations where task-specific data is scarce. This challenge arises from the disparity between pre-training tasks, such as masked language modeling, and the subsequent fine-tuning tasks encompassing classification and regression. This divergence hampers the seamless adaptation of pre-trained language models (PLMs) to downstream tasks.

\begin{figure}[!t]
	\centering
	\includegraphics[width=1\linewidth]{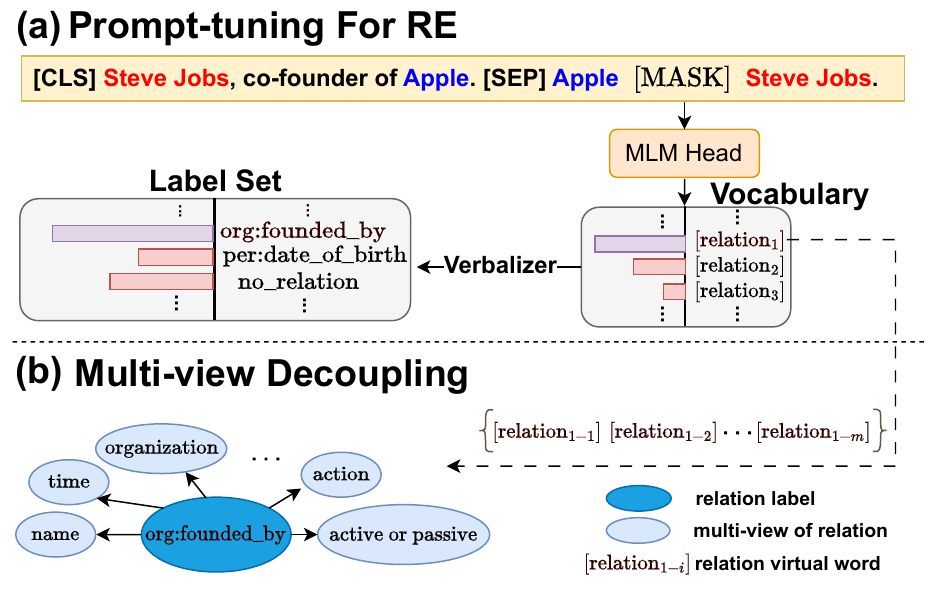}
	\caption{(a) An example of prompt-tuning for RE. Red-colored words indicate the subject, while blue-colored words indicate the object. (b) The concept of multi-view decoupling attempts to encompass various aspects of a relation using multiple relation representations.
 }
	\label{fig:intro}
\end{figure}

Recently, prompt tuning has emerged as a promising direction for facilitating few-shot learning, which effectively bridges the gap between the pre-training and the downstream task~\cite{gao2021making,jin2023instance}. Conceptually, prompt-tuning involves template and verbalizer engineering, aiming to discover optimal templates and answer spaces. For example, as shown in Figure~\ref{fig:intro} (a), given a sentence ``Steve Jobs, co-founder of Apple" for relation extraction, the text will first be enveloped with relation-specific templates, namely transforming the original relation extraction task into a relation-oriented cloze-style task. Subsequently, the PLM will predict words in the vocabulary to fill in the [MASK] position, and these predicted words are finally mapped to corresponding labels through a verbalizer. 
In this example, the filled word ``$[\text{relation}_1]$" (e.g., ``founded") can be linked to the label ``org:founded\_by" through the verbalizer. 
However, for complex relation representations, such as ``per: country\_of\_birth" and ``org: city\_of\_headquarters," obtaining suitable vocabulary labels is much more challenging. To address this issue, previous work \cite{han2022ptr} applies logic rules to decompose complex relations into descriptions related to the subject and object entity types. Some works
 construct virtual words for each relation (a trainable  ``$[\text{relation}_1]$") to substitute the corresponding answer space of the complex relation~\cite{chen2022knowprompt,chen2022relation}.
This paradigm focuses on optimizing the relation representation space and demands PLMs to learn representations for words that are not present in the vocabulary. 
However, in extremely low-resource scenarios, such as one-shot RE, building robust relation representations with this paradigm is difficult, thus leading to a performance drop.

To mitigate the above issue, in this paper,  we introduce Multi-view Relation Extraction (MVRE), which improves low-resource prompt-based relation representations with a multi-view decoupling framework. As illustrated in Figure~\ref{fig:intro} (b), considering that relations may contain multiple dimensions of information, for instance, ``org:founded\_by" may entail details about organizations, people's names, time, the action of founding, and so on. According to theoretical analysis, being limited to a single vector representation, the model may face the upper boundary of representation capacity and fail to construct robust representations in low-resource scenarios. Therefore, we propose to optimize the latent space by decoupling it into a joint optimization of multi-view relation representations, thereby maximizing the likelihood during relation inference. By sampling a greater number of relation representations, as denoted ``$[\text{relation}_{1-i}]$" in Figure~\ref{fig:intro} (b)), we promote the learned latent space to include more kinds of information about the corresponding relation. In detail, we achieve this decoupling process by disassembling the virtual words into multiple components and predicting these components through successive [MASK] tokens. Furthermore, we introduce a Global-Local loss and Dynamic Initialization approach to optimize the process of relation representations by constraining semantic information of relations. We evaluate MVRE on three
relation extraction datasets. Experimental results demonstrate that our
method significantly outperforms previous approaches. To sum up, our main contributions are as follows:

\begin{itemize}

\item To the best of our knowledge, this paper presents the first attempt to improve low-resource prompt-based relation representations with multi-view decoupling learning. In this way, the PLM can be comprehensively utilized for generating robust relation representations from limited data. 
\item To optimize the learning process of multi-view relation representations, we introduce the Global-Local Loss and Dynamic Initialization to impose semantic constraints between virtual relation words.
\item We conduct extensive experiments on three datasets and our proposed MVRE can achieve state-of-the-art performance in low-resource scenarios.

\end{itemize}
%


\section{Background and Related Work}

\subsection{Prompt-Tuning for RE}
Inspired by the ``in context learning" proposed in GPT-3~\cite{brown2020language}, the approach of stimulating model knowledge through a few prompts has recently attracted increasing attention.
In text classification tasks, significant performance gains can be achieved by designing a tailored prompt for a specific task, particularly in few-shot scenarios~\cite{schick2021exploiting,gao2021making}. In order to alleviate the labor-intensive process of manual prompt creation, there has been extensive exploration into automatic searches for discrete prompts~\cite{schick2020automatically,wang2022automatic} and continuous prompts~\cite{huang2022fpt,gu2022ppt}.


For RE with prompt-tuning, a template function can be defined in the following format: $T(x) = x : w_s : [\text{MASK}] : w_o$, where ``:" signifies the operation of concatenation. By employing this template function, the instance $x$ is modified to incorporate the entity pair $(w_s, w_o)$, resulting in the formation of $x_{prompt}=T(x)$. In this process, $x_{prompt}$ is the corresponding input of model $M$ with a [MASK] token in it.
Here, $Y$ refers to the label words set, and $\mathcal{V}$ donates the relation set within the prompt-tuning framework. A verbalizer $v$ is a mapping function $v: Y \longrightarrow \mathcal{V}$, establishing a connection between the relation set and the label word set, where $v(y)$ means label words corresponding to label $y$. 
The probability distribution over
the relation set is calculated as:
\begin{equation}
    p(y|x)=p_{M}([\text{MASK}]=v(y)|T(x))  \label{eq:prompt}
\end{equation}
 In this way, the RE problem can be transferred into a masked language modeling problem by filling the [\text{MASK}] token in the input. 


However, for relation extraction, the complexity and diversity of relations pose challenges in employing these methods to discover suitable templates and answer spaces.
\citet{han2022ptr} propose prompt-tuning methods for RE by applying logic rules to construct hierarchical prompts.
\citet{lu2022summarization} make prompts for each relation and converts RE into a generative summarization problem. These works translate the prediction of a relation into predicting a specific sentence, which to some extent addresses the complexity of relations. However, summarizing the intricate information of a relation using these words remains challenging.


\subsection{Virtual Relation Word}
\citet{chen2022knowprompt} introduce virtual relation words and leverage prompt-tuning for RE by injecting semantics of relations and entity types. \citet{chen2022relation} propose retrieval-enhanced prompt-tuning by incorporating retrieval of representations obtained through prompt-tuning. These studies devise virtual words for each relation in prompt-tuning, circumventing the need to search for complex answer spaces~\cite{liu2023pre}.

The corresponding verbalizer $v^{*}$ for this approach function as $v^{*}: Y \longrightarrow \mathcal{V}^{*}$, where $\mathcal{V}^{*}=\{\mathcal{V}, \mathcal{V}^{Y}\}$,$|Y|=|\mathcal{V}^{Y}|$,$v^{*}(y)\in \mathcal{V}^{Y}$,$y\in Y$. The $\mathcal{V}^{Y}$ corresponds to virtual relational words, representing the set of words created for each relation.
The acquisition of this virtual word for a relation is equivalent to obtaining a latent space representation for that relation. As the relation virtual words do not exist in the pre-trained model's vocabulary, ensuring robust representations often requires a sufficient amount of data or semantic constraints to the prompt-based instance representation~\cite{chen2022relation}.


Given an instance $x$, the prompt-based instance representation $h^x$ can be computed by leveraging the output embedding of the ``$[\text{MASK}]$" token of the last layer of the underlying PLM:
\begin{equation}
    h^x = M(T(x))_{[\text{MASK}]} \label{eq:mask_emb}
\end{equation}
The prompt-based instance representation $h^x$ can capture the relation corresponding to the instance $x$, and ultimately, through the ``MLM head", derive the classification probabilities for the respective virtual relation word~\cite{chen2022knowprompt,chen2022relation}. Most of these approaches confine a complex relation to a single prompt-based vector,
which limits the learning of relation latent space in low-resource scenarios.

\section{Method}
\begin{figure*}[!t]
	\centering
	\includegraphics[width=0.99\linewidth]{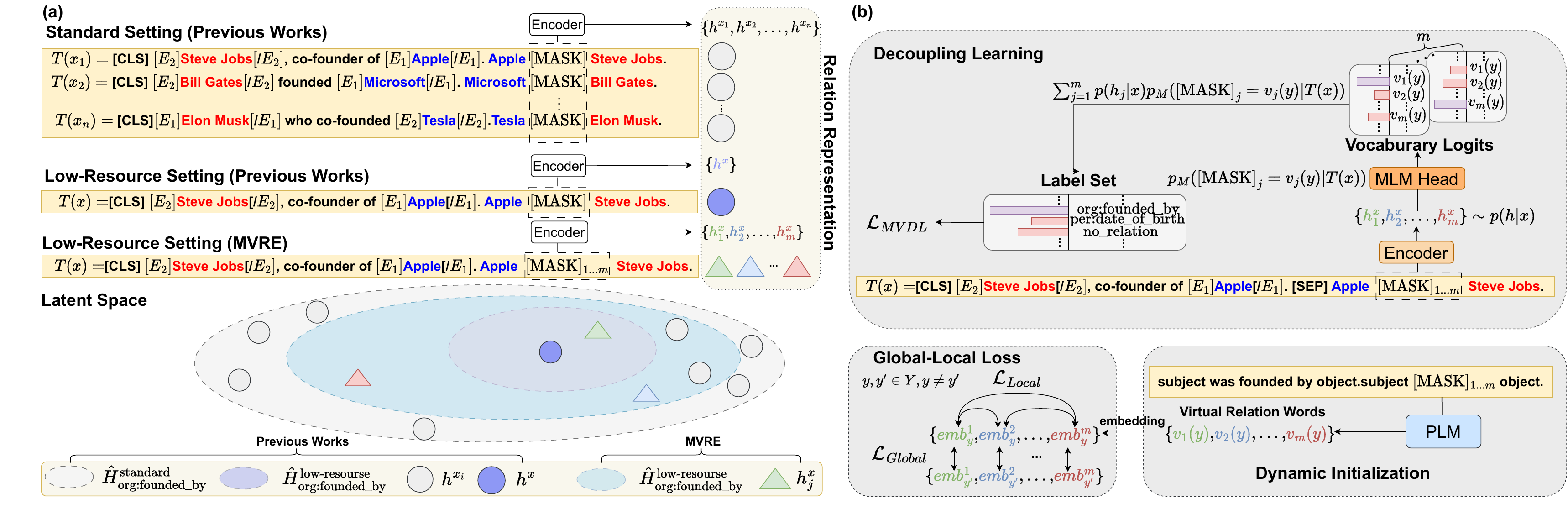}
	\caption{(a) An illustrative comparison of the relation latent space learning process between MVRE and previous prompt-based works. We employ multi-view relation representations to cover a broader latent space in low-resource scenarios. (b) The MVRE framework incorporates Multi-view Decoupling Learning,  Global-Local Loss and Dynamic Initialization processes.
 }
	\label{fig:framework}
\end{figure*}
\subsection{Preliminaries}

Formally, a RE dataset can be denoted as $D = \{X, Y\}$, where $X$ is the set of examples and $Y$ is the set of relation labels. For each example $x=\{w_1,w_2,w_s,...,w_o,...,w_n\}$, the goal of RE is to predict the relation
$y \in Y$ between subject entity $w_s$ and object entity $w_o$ (since an entity may have multiple tokens, we simply utilize $w_s$ and $w_o$ to
represent all entities briefly).

\subsubsection{Previous prompt-tuning in Standard Scenario}

In the prompt-based instance learning for relations, it is assumed that for each class $y_i$, we learn a corresponding latent space representation $H_{y_i}$ such that $F^{-1}(y_i) = H_{y_i}$, where $F$ denotes the function mapping between labels and representations.
In the case of a standard scenario, where all available data can be used, the model minimizes the following loss function:
\begin{equation}
\begin{aligned}
&\mathbb{E}_{x\sim \mathcal{X}}[-\log p(y|x)] = -\frac{1}{N}\sum_{i=1}^N \log p(y_i,H_{y_i}|x_i)
\label{eq:all}
\end{aligned}
\end{equation}
where $N$ represents the total data volume across all classes. In this process, focusing solely on a specific relation $y_e$, the learned latent space representation $\hat H^{\text{standard}}_{y_e}$ for class $y_e$ satisfies $F(h^{x^e_i})=y_e$, where $1\leq i \leq \#y_e$ and$(x^e_i, y_e) \in (\mathcal{X,Y})$. Here, $\#y_e$ represents the number of instances in the data with the label $y_e$.
The process of obtaining $\hat H^{\text{standard}}_{y_e}$ is akin to optimizing the following expression:
\begin{equation}
\begin{aligned}
    & \min_{\theta}\sum_{(x^e_i,y_e)\in (X,Y)}sim(H_{y_e},F^{-1}(y_e,\theta))
    \label{eq:H_standard}
\end{aligned}
\end{equation}
where ``$sim$" represents the degree of similarity between the latent space representations. However, in low-resource scenarios, the value of $\#y_e$ can constrain the optimization effectiveness of Eq~\ref{eq:H_standard}.

\subsection{Multi-view Decoupling Learning}

Therefore, we assume that in the process of learning the complex relation latent space $H_{y_i}$, it is feasible to decompose this space into multiple perspectives and learn from various viewpoints. Consequently, we consider the learning process for single data pair $(x_i,y_i)$ as follows:
\begin{equation}
\begin{aligned}
    &p(y_i,H_{y_i}|x_i)=\sum_{h} p(y_i,h|x_i) \\
    &=\sum_{h} p(y_i|x_i,h)p(h|x_i)\\
    &=\mathbb{E}_{h\sim p(h|x_i)}p(y_i|x_i,h) \label{eq:multi_h}
\end{aligned}
\end{equation}
where $h$ represents a perspective in which the relation $y_i$ is decomposed, we transform the learning of relations into the process of learning each relation's various perspectives. Ultimately, we merge the information from multiple perspectives to optimize the relation inference process.

Similar to Eq~\ref{eq:H_standard}, when there is only one pair of data for a given relation, the learning of its latent space is as follows:
\begin{equation}
\begin{aligned}
    &\min_{\theta}\sum_{(x^e,y_e)\in (\mathcal{X,Y}), y_e^j\in y_e}sim(H_{y_e},F^{-1}(y_e^j,\theta))\label{eq:H_1shot}
\end{aligned}
\end{equation}

In this process, the learned latent space representation $\hat H^{\text{1-shot}}_{y_e}$ for class $y_e$ satisfies $F(h^{x^e}_j)=y_e$, where $1\leq j \leq m$ and $(x^e, y_e) \in (\mathcal{X,Y})$. Here, $m$ represents the number of decomposed perspectives for the relation $y_e$.

\subsubsection{Sampling of Relation Latent Space} \label{text:sample}

Under normal circumstances, the latent space learned in a low-resource setting tends to be inferior compared to the standard scenario, resembling $sim(\hat H^{\text{1-shot}}_{y_e},H_{y_e}) \geq sim(\hat H^{\text{standard}}_{y_e},H_{y_e})$. Hence, as can be seen in Figure~\ref{fig:intro} (a), our objective is for the low-resource acquired latent space to closely resemble that learned in the standard scenario, as  $E(\hat H^{\text{1-shot}}_{y_e}) \sim  E(\hat H^{\text{standard}}_{y_e})$. Combining Eq~\ref{eq:H_standard} and Eq~\ref{eq:H_1shot}, the representation set $\{h^{x^e}_j|1\leq j \leq m\}$ we acquire needs to resemble the representation set $\{h^{x^e_i}|1\leq i \leq \#y_e\}$ obtained under standard conditions. 
This highlights the necessity of sampling a substantial number of $h^{x^e}_j(m \geq 1)$ instances with similar distribution to ensure alignment of the obtained relation latent space with that in standard scenarios. The value of $m$ will be discussed in the experimental section.

According to the Eq~\ref{eq:mask_emb}, $h$ is determined by the parameters of model $M$, the structure of template $T$, and the expression ``$\text{[MASK]}=v(y_i)$":
\begin{equation}
\begin{aligned}
    &p(y_i|x_i,h^{x_i}) = p(y_i|x_i,M(T(x_i))_{[\text{MASK}]})\\&=p_{M}([\text{MASK}]={v(y_i)}|T(x_i)) \label{eq:7}
\end{aligned}
\end{equation}

To ensure a consistent interpretation of $h^{x_i}$ obtained from single data pair, while simultaneously covering various perspectives of a relation, we sample $h^{x_i}$ based on the expression ``$\text{[MASK]} = v(y_i)$". Specifically, we expand the token ``$\text{[MASK]}$" into multiple contiguous tokens within the template, each ``$\text{[MASK]}$" corresponds to as follows:
\begin{equation}
    T(x)=x:[sub]:[\text{MASK}]_{\{1...m\}}:[obj]
    \label{eq:template}
\end{equation}
the sampling method for $h^{x_i}_{j}$ is as follows, $h^{x_i}_{j} = M(T(x))_{[\text{MASK}]_j}$. It's important to note that a relation in text can be represented by a continuous segment of text. Therefore, this approach has the potential to capture multi-view representations of a relation.

Based on our sampling method for latent space representation, we derive the probability distribution of $y_i$ as follows:
\begin{equation}
\begin{aligned}
    &p(y_i|x_i,h^{x_i}_j)=p_{M}([\text{MASK}]_j={v_j(y_i)}|T(x_i))
\end{aligned}
\end{equation}
Due to the challenge of finding suitable words in the vocabulary to match different perspectives of a relation, we introduce $m$ new multi-view virtual relation words, denoted as $v_j(y)$, for each relation $y_i$. Combining Eq~\ref{eq:multi_h}, the final loss function $\mathcal{L}_{\text{MVDL}}(x_i,y_i)$ that the model needs to minimize is as follows:
\begin{equation}
    \sum_{j=1}^m\Big[-\log \Big( p(h_j^{x_i}|x_i)p_{M}([\text{MASK}]_j={v_j(y_i)}|T(x_i)\Big)\Big]
\end{equation}
 Here, we employ a matrix $W_h$ to learn the posterior probability of $h_j^{x_i}$, the formula is as follows $p(h_j^{x_i}|x_i)=\frac{\sigma(W_h^\mathrm{T} h_j^{x_i})}{\sum_{k=1}^m \sigma(W_h^\mathrm{T} h_k^{x_i})}$, where $\sigma$ represents the sigmoid function.

When considering all the data, the loss function is given by:
\begin{equation}
    \mathcal{L}_{\text{MVDL}}=\sum_{(x_i,y_i)\in (X,Y)}\mathcal{L}_{\text{MVDL}}(x_i,y_i)\label{eq:mvre_loss}
\end{equation}

\subsubsection{Global-Local Loss}
The contrastive learning methods to enhance representation learning have been employed in many previous works~\cite{gao2021simcse,zhang2022multi}.
To encourage better alignment of multi-view virtual relation words $v_j(y)$ with diverse semantic meanings, we introduce the Global-Local Loss(referred to as ``GL") to optimize the learning process of multi-view relation virtual words. The Local Loss encourages virtual words representing the same relation to focus on similar information, while the Global Loss ensures that virtual words representing different relations emphasize distinct aspects. Their expressions are as follows:

\begin{equation}
\begin{aligned}
    &\mathcal{L}_{\text{Local}}=-\frac{1}{|Y|m^2}\sum_{r\in Y}\left[\sum_{i,j\in[1,m]} sim(emb_r^i,emb_r^j)\right]\\
    &\mathcal{L}_{\text{Global}}=\frac{1}{|Y|^2m}\sum_{i=1}^m \left[\sum_{ru,rv \in \mathcal{R}}  sim(emb_{ru}^i,emb_{rv}^i)\right]
\end{aligned} 
\end{equation}
where $sim(x,y)=cos(\frac{x}{||x||},\frac{y}{||y||})$, $emb_r^i$ denotes the embedding of the virtual word for relation $v_i(r)$.

Finally, the loss function of MVRE is as follows:
\begin{equation}
    \mathcal{L}_{\text{MVRE}}=\mathcal{L}_{\text{MVDL}}+\alpha * \mathcal{L}_{\text{Local}}+\beta * \mathcal{L}_{\text{Global}}
    \label{eq:final_loss}
\end{equation}
where $\alpha$ and $\beta$ are hyperparameters. The framework of MVRE is illustrated in Figure~\ref{fig:framework} (b).

\subsubsection{Dynamic Initialization}


The virtual word for a relation also involves learning a new word that does not exist in the original vocabulary.
Therefore, efficient initialization is crucial for achieving desirable results in this process. However, in MVRE, it is essential to have meaningful initialization methods that consider the actual positions of each virtual word in the text.

We introduce Dynamic Initialization (referred to as ``DI"), which leverages the PLM's cloze-style capability to identify appropriate initialization tokens for relation-representing virtual words. Specifically, we first create a manual template for each relation and insert a prompt after it (The manual template for each relation can be found in the appendix C). Then, we employ the model to find the token with the highest probability, which serves as the initialization token for the respective virtual word. To enhance the construction of relation information, we incorporate the entity information corresponding to the label itself. This knowledge is not involved in the model's training process and is similar to prompts, as it leverages the inherent abilities of the model, thus preserving the characteristics of low-resource scenarios.

To mitigate the potential generation of irrelevant tokens during dynamic initialization, particularly with larger $m$ values, we merge the static and dynamic initialization techniques. Inspired by ~\citet{chen2022knowprompt}, we introduce \textbf{Static Initialization} (referred to as ``SI"), where words for initialization are derived from the labels corresponding to each relation. We integrate the two methods by averaging the tokens' embedding obtained from static and dynamic initialization.

\section{Experiments}

\begin{table}[]
\centering
\renewcommand\arraystretch{1.5}
\begin{tabular}{l|c|c|c|c} 
\hline
Dataset  & Train & Dev  & Test & Relation \\ \hline
SemEval  & 6,507   & 1,493  & 2,717  & 19    \\
TACRED   & 68,124  & 22,631 & 15,509 & 42    \\
TACREV   & 68,124  & 22,631 & 15,509 & 42   \\ \hline
\end{tabular}
\caption{The statistics of different RE datasets.}
\label{tab:table-stat}
\end{table}

\subsection{Datasets}

For comprehensive experiments, we conduct experiments on three RE datasets: SemEval 2010 Task 8 (SemEval) \cite{hendrickx2010semeval},
TACRED \cite{zhang2017position}, and TACRED-Revisit (TACREV) \cite{alt2020tacred}. Here we briefly describe them below. The detailed statistics are provided in Table~\ref{tab:table-stat}.

\noindent \textbf{SemEval} is a traditional dataset in relation extraction that does not provide entity types. It covers 9 relations with two directions and one special relation ``Other".

\noindent \textbf{TACRED} is one large-scale sentence-level relation extraction dataset drawn from the yearly TACKBP4 challenge, which contains 41 common relation types and a special ``no relation" type.

\noindent \textbf{TACREV} builds on the original TACRED dataset. They find out and correct the errors in the original development set and test set of TACRED, while the training set is left intact.TACREV and TACRED share the same set of relation types.

\begin{table*}[]
\centering
\resizebox{\linewidth}{!}{
\begin{tabular}{l|lll|lll|lll}
\hline
\multicolumn{1}{c|}{\multirow{2}{*}{Model}} & \multicolumn{3}{c|}{SemEval}                                       & \multicolumn{3}{c|}{TACRED} & \multicolumn{3}{c}{TACREV} \\ \cline{2-10} 
\multicolumn{1}{c|}{}                       & \multicolumn{1}{c}{K=1}                         & \multicolumn{1}{c}{K=5}                         & \multicolumn{1}{c|}{K=16}   & \multicolumn{1}{c}{K=1}     & \multicolumn{1}{c}{K=5}     & \multicolumn{1}{c|}{K=16}    & \multicolumn{1}{c}{K=1}     & \multicolumn{1}{c}{K=5}    & \multicolumn{1}{c}{K=16}    \\ \hline

\multicolumn{3}{l}{\textit{Compared Methods}} \\\hline
$\text{FINE-TUNING}$                                      & 18.5{\scriptsize($\pm 1.4$)} & 41.5{\scriptsize($\pm 2.3$)} & 66.1{\scriptsize($\pm 0.4$)} & 7.6{\scriptsize($\pm 3.0$)}      & 16.6{\scriptsize($\pm 2.1$)}       & 26.8{\scriptsize($\pm 1.8$)}       & 7.2{\scriptsize($\pm 1.4$)}      & 16.3{\scriptsize($\pm 2.1$)}      & 25.8{\scriptsize($\pm 1.2$)}       \\
$\text{GDPN}_{\text{ET}}$                                      & 10.3{\scriptsize($\pm 2.5$)} & 42.7{\scriptsize($\pm 2.0$)} & 67.5{\scriptsize($\pm 0.8$)} & 4.2{\scriptsize($\pm 3.8$)}      & 15.5{\scriptsize($\pm 2.3$)}       & 28.0{\scriptsize($\pm 1.8$)}       & 5.1{\scriptsize($\pm 2.4$)}      & 17.8{\scriptsize($\pm 2.4$)}      & 26.4{\scriptsize($\pm 1.2$)}       \\
PTR                                         & 14.7{\scriptsize($\pm 1.1$)} & 53.9{\scriptsize($\pm$1.9)} & 80.6{\scriptsize($\pm$1.2)} & 8.6{\scriptsize($\pm 2.5$)}      & 24.9{\scriptsize($\pm 3.1$)}       & 30.7{\scriptsize($\pm 2.0$)}       & 9.4{\scriptsize($\pm 0.7$)}       & 26.9{\scriptsize($\pm 1.5$)}      & 31.4{\scriptsize($\pm 0.3$)}       \\
KnowPrompt                                  & 28.6{\scriptsize($\pm 6.2$)} & 66.1{\scriptsize($\pm 8.6$)} & 80.9{\scriptsize($\pm 1.6$)} & 17.6{\scriptsize($\pm 1.8$)}      & 28.8{\scriptsize($\pm 2.0$)}       & 34.7{\scriptsize($\pm 1.8$)}       & 17.8{\scriptsize(($\pm 2.2$)}       & 30.4{\scriptsize($\pm 0.5$)}      & 33.2{\scriptsize($\pm 1.4$)}       \\
RetrievalRE                                 & 33.3{\scriptsize($\pm 1.6$)} & 69.7{\scriptsize($\pm 1.7$)} & 81.8{\scriptsize($\pm 1.0$)} & 19.5{\scriptsize($\pm 1.5$)}      & 30.7{\scriptsize($\pm 1.7$)}       & \textbf{36.1}{\scriptsize($\pm 1.2$)}       & 18.7{\scriptsize($\pm 1.8$)}       & 30.6{\scriptsize($\pm 0.2$)}      & \textbf{35.3}{\scriptsize($\pm 0.3$)}       \\ \hline
\multicolumn{3}{l}{\textit{Ours}} \\
\hline
MVRE(w/o GL\&DL)                                        & 35.3{\scriptsize($\pm 4.6$)}  &  74.6{\scriptsize($\pm 1.7$)}    & 81.3{\scriptsize($\pm 1.4$)}       &  21.0{\scriptsize($\pm 2.1$)}       & 31.4{\scriptsize($\pm 1.0$)}       &  32.9{\scriptsize($\pm 2.5$)}       &  20.2{\scriptsize($\pm 0.7$)}       & \textbf{31.0}{\scriptsize($\pm 1.1$)}        &  34.1{\scriptsize($\pm 2.1$)} \\
MVRE                                        & \textbf{54.6}{\scriptsize($\pm 2.8$)}  &  \textbf{77.6}{\scriptsize($\pm 3.6$)}    & \textbf{82.5}{\scriptsize($\pm 0.8$)}       &  \textbf{21.2}{\scriptsize($\pm 2.2$)}       & \textbf{32.4}{\scriptsize($\pm 1.2$)}       &  34.8{\scriptsize($\pm 0.8$)}       &  \textbf{20.5}{\scriptsize($\pm 1.9$)}       & \textbf{31.0}{\scriptsize($\pm 1.4$)}        &  34.3{\scriptsize($\pm 1.1$)} \\
 \hline
\end{tabular}
}
\caption{Performance of RE models in the low-resource setting. We report the mean and standard deviation performance of micro $F1$ scores (\%) over 5 different splits. The best numbers are highlighted in each column.}
\label{tab:table-few-shot}
\end{table*}

\subsection{Compared Methods}

To evaluate our proposed MVRE, we compare with the following methods: 
~(1) \textbf{FINE-TUNING} employs a conventional fine-tuning approach for PLMs to relation extraction.
~(2) \textbf{$\text{GDPN}_{\text{ET}}$} utilizes the multi-view graph for relation extraction~\cite{xue2021gdpnet} 
~(3) \textbf{PTR}~\cite{han2022ptr} propose prompt-tuning methods for RE by applying logic rules to partition relations into sub-prompts;~(4) \textbf{KnowPrompt}~\cite{chen2022knowprompt} utilize virtual relation
word to prompt-tuning;
~(5) \textbf{RetrieveRE}~\cite{chen2022relation} employ retrieval to enhance prompt-tuning.

\subsection{Implementation Details}

We utilize Roberta-large for all experiments to make a fair comparison. For test metrics, we use micro $F_1$ scores of RE as
the primary metric to evaluate models, considering that $F_1$ scores
can assess the overall performance of precision and recall. 
More detailed settings can be found in the Appendix A. 

\textbf{Low-resource Setting.} we adopt the same setting as RetrievalRE~\cite{chen2022relation} and perform experiments using 1-, 5-, and 16-shot scenarios to evaluate the performance of our approach in extremely low-resource situations. To avoid randomness, we employ a fixed set of seeds to randomly sample data five times and record the average performance and variance. During the sampling process, we select $k$ instances for each relation label from the original training sets to compose the few-shot training sets.

\textbf{Standard Setting.} In the standard setting, we leverage full trainsets to conduct experiments and compare with previous prompt-tuning methods, including PTR, KnowPrompt, and RetrievalRE.

\begin{table*}[ht]
\centering
\resizebox{\linewidth}{!}{
\begin{tabular}{lll}
\hline
\multicolumn{3}{l}{\begin{tabular}[c]{@{}l@{}}x=The National Congress of American Indians was founded in 1944 in response to the implementation of \\ assimilation policies  on tribes by the federal government.\\ {[}sub{]}=National Congress of American Indians {[}obj{]}=1944\end{tabular}}                                                                                                                                                                                                                                                                                           \\ \hline
\multicolumn{1}{l|}{m}                                                         & \multicolumn{1}{l|}{top-1 token; T(x)=x {[}sub{]} {[}MASK{]}*m {[}obj{]}}                                                                                                                                                                                  & top-1 token; T(x)=x {[}obj{]} {[}MASK{]}*m {[}sub{]}                                                                                                                                                                                                   \\ \hline
\multicolumn{1}{c|}{\begin{tabular}[c]{@{}c@{}}1\\ 2\\ 3\\ 4\\ 5\end{tabular}} & \multicolumn{1}{l|}{\begin{tabular}[c]{@{}l@{}}in(0.42)\\ founded(0.48)  in(0.70)\\ \textbf{was(0.87)  founded(0.92)  in(0.93)}\\ was(0.46)  was(0.16)  founded(0.19)  in(0.55)\\ was(0.44)  founded(0.31)  in(0.29)  founded(0.03)  ,(0.74)\end{tabular}}     & \begin{tabular}[c]{@{}l@{}}.(0.31)\\ .(0.18) The(0.19)\\ .(0.08) of(0.59) the(0.48)\\ \textless{}/s\textgreater{}(0.07) of(0.03)  of(0.53) the(0.55)\\ \textbf{\textless{}/s\textgreater{}(0.09) the(0.05) founding(0.09) of(0.63) the(0.70)}\end{tabular} \\ \hline
\multicolumn{3}{l}{\begin{tabular}[c]{@{}l@{}}x=The series reflected on the changes that had taken place in Ireland since the 1960s.\\ {[}sub{]}=series {[}obj{]}=changes\end{tabular}}                                                                                                                                                                                                                                                                                                                                                                                                    \\ \hline
\multicolumn{1}{l|}{m}                                                         & \multicolumn{1}{l|}{top-1 token; T(x)=x {[}sub{]} {[}MASK{]}*m {[}obj{]}}                                                                                                                                                                                  & top-1 token; T(x)=x {[}obj{]} {[}MASK{]}*m {[}sub{]}                                                                                                                                                                                                   \\ \hline
\multicolumn{1}{c|}{\begin{tabular}[c]{@{}c@{}}1\\ 2\\ 3\\ 4\\ 5\end{tabular}} & \multicolumn{1}{l|}{\begin{tabular}[c]{@{}l@{}}on(0.20)\\ reflected(0.24) those(0.34)\\ \textbf{reflected(0.69) on(0.87) those(0.40)}\\ reflected(0.15) on(0.10) on(0.27) those(0.41)\\ reflected(0.08) the(0.12) some(0.06) of(0.22) those(0.43)\end{tabular}} & \begin{tabular}[c]{@{}l@{}}the(0.41)\\ in(0.53) the(0.83)\\ to(0.06) in(0.18) the(0.64)\\ are(0.12) reflected(0.05) throughout(0.35) the(0.69)\\ \textbf{that(0.10) been(0.08) reflected(0.07) in(0.30) the(0.69)}\end{tabular}                            \\ \hline
\end{tabular}
}
\caption{Case study of Dynamic Initialization. Each line represents the top-1 token generated for each [MASK] and its corresponding probability when the number of [MASK] is $m$. We highlight the parts that represent the relation more accurately}
\label{tab:my-table-init}
\end{table*}
\begin{table}[]
\centering
\resizebox{\linewidth}{!}{
\begin{tabular}{llllllll}
\hline
\multicolumn{1}{l|}{Method}          & GL        & SI        & \multicolumn{1}{l|}{DI}        & K=1           & K=5           & K=16          & Full          \\ \hline
\multicolumn{1}{l|}{\multirow{6}{*}{MVRE}} & \checkmark & \checkmark & \multicolumn{1}{l|}{\checkmark} & \textbf{54.6} & \textbf{77.6} & \textbf{82.5} & 90.2          \\
\multicolumn{1}{l|}{}                      &           & \checkmark & \multicolumn{1}{l|}{\checkmark} & \textbf{54.6} & 77.1          & 82.1          & 89.3          \\
\multicolumn{1}{l|}{}                      & \checkmark &           & \multicolumn{1}{l|}{\checkmark} & 44.9          & 74.1          & 82.4          & 89.8          \\
\multicolumn{1}{l|}{}                      &           &           & \multicolumn{1}{l|}{\checkmark} & 43.3          & 73.1          & \textbf{82.5} & 89.5          \\
\multicolumn{1}{l|}{}                      & \checkmark & \checkmark & \multicolumn{1}{l|}{}          & 37.5          & 72.9          & 81.5          & 89.5          \\
\multicolumn{1}{l|}{}                      &           & \checkmark & \multicolumn{1}{l|}{}          & 35.3          & 74.6          & 81.3          & 89.9          \\ \hline
\multicolumn{8}{l}{Prompt-tuning Pre-trained Model(For Reference)}                                                                                                  \\ \hline
\multicolumn{1}{l|}{PTR}                   &           &           & \multicolumn{1}{l|}{}          & 14.7          & 53.9          & 80.6          & 89.9          \\
\multicolumn{1}{l|}{KnowPrompt}            &           & \checkmark & \multicolumn{1}{l|}{}          & 28.6          & 66.1          & 80.9          & 90.2          \\
\multicolumn{1}{l|}{RetrievalRE}           &           & \checkmark & \multicolumn{1}{l|}{}          & 33.3          & 69.7          & 81.8          & \textbf{90.4} \\ \hline
\end{tabular}
}
\caption{Ablation Study on SemEval: Investigating the Impact of Global-Local Loss (GL), Static Initialization (SI), and Dynamic Initialization (DI). The "Full" column indicates the results under the standard setting.}
\label{tab:my-table-ablation}
\end{table}

\subsection{Low-Resource Results}
We present our results on low-resource settings in Table~\ref{tab:table-few-shot}.
Notably, across all datasets, our MVRE consistently outperforms all previous prompt-tuning models. Particularly remarkable is the substantial improvement in the 1-shot scenario, with gains of 63.9\%, 8.7\%, and 9.6\% over RetrievalRE in SemEval, TACRED, and TACREV respectively. When $k$ is set to 5 or 16, the magnitude of improvement decreases. In the TACRED and TACREV datasets, when $k$ is set to 16, there's a slight decrease compared to the retrieval-enhanced RetrievalRE. However, overall, the performance remains better than KnowPrompt, a fellow one-stage prompt-tuning method similar to ours. Similar to previous works~\cite{chen2022knowprompt,chen2022relation}, the comparison of performance between fine-tuning-based methods(FINE-TUNING, $\text{GDPN}_{\text{ET}}$) and MVRE demonstrates the superiority of prompt-based methods in low-resource settings. 

 It's noteworthy that our method doesn't exhibit the same significant improvements in TACRED and TACREV as observed in SemEval. Our speculation is attributed to two reasons: (1) In TACRED and TACREV, the high proportion of ``other" relations (78\% in TACRED/V, 17\% in SemEval) can make it challenging to categorize relations as ``other" in the low-resource scenario. (2) There are more similar relations than SemEval, such as ``org:city\_of\_headquarters" and ``org:stateorprovince\_of\_headquarters", making it more difficult to distinguish them in low-resource scenarios. 


\subsection{Ablation Study}
To prove the effects of the components of MVRE, including Global-Local Loss(GL), Dynamic Initialization(DI), and Static Initialization(SI), we conduct the ablation study on SemEval and present the results in Table~\ref{tab:my-table-ablation}.
Additionally, we present the results under the standard setting in Table ~\ref{tab:my-table-ablation}.
\subsubsection{Standard Results}
Under the full data scenario, MVRE and KnowPrompt yield equivalent results, indicating that our approach remains applicable and does not compromise model performance when enough data is available. 
\subsubsection{Global-Local Loss}
As observed in Table~\ref{tab:my-table-ablation}, the incorporation of the Global-Local Loss(GL) consistently yields improvements across various scenarios, resulting in an enhancement of the relation F1 score by 0.5, 0.4, and 0.5 in the 5-shot, 16-shot, and standard settings, respectively. This phenomenon demonstrates that constraining the semantics of virtual relation words' embedding through a comparative method can optimize the representation of multi-perspective relations.
\subsubsection{The Initialization of Virtual Relation Words}
We also conduct an ablation study to validate the effectiveness of the initialization of relation virtual words. Previous studies have revealed that achieving satisfactory relation representations with random initialization is challenging~\cite{chen2022knowprompt}.
Hence, to ensure model performance, it is essential to use either Static Initialization(SI) or Dynamic Initialization(DI) during the experiment. When both are employed simultaneously, their corresponding tokens' embedding is averaged to integrate these two methods. Table~\ref{tab:my-table-ablation} demonstrates that adopting Dynamic Initialization leads to a significant enhancement in model performance compared to Static Initialization. Furthermore, combining both initialization methods also yields substantial improvements.

\subsubsection{Effect of m Number of [MASK]}
\begin{figure}[!t]
	\centering
	\includegraphics[width=0.99\linewidth]{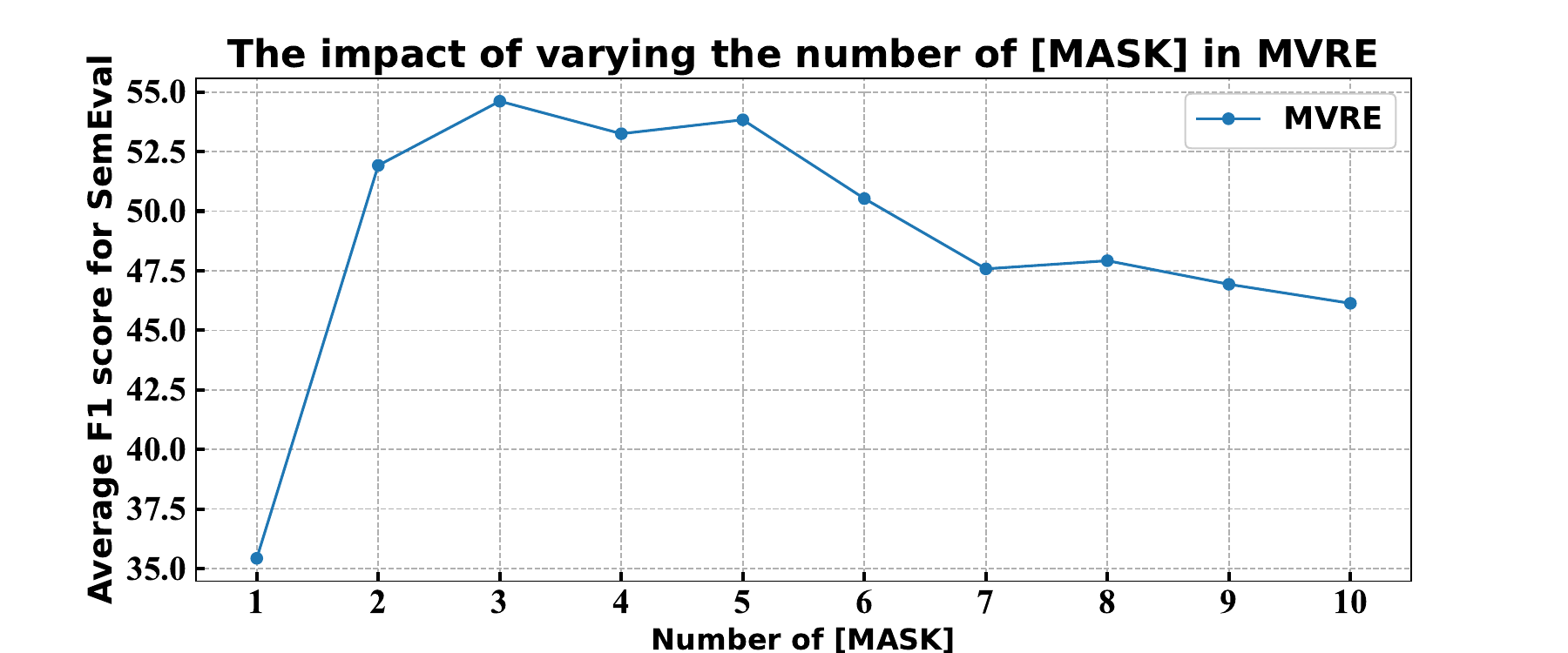}
	\caption{Effect of the number of [MASK] on MVRE. }
	\label{fig:many_mask}
\end{figure}

Due to the introduction of noise when inserting ``[MASK]" and further, the efficiency of decoupling learning presents significant challenges. Therefore, simply increasing the number of ``[MASK]" tokens cannot enhance performance in low-resource scenarios. As shown in Figure~\ref{fig:many_mask}, we conduct experiments to investigate the impact of varying quantities of ``[MASK]" tokens on relation extraction effectiveness, aiming to identify the optimal value for $m$. The performance of the model shows a trend of initially increasing and then decreasing as the value of $m$ increases. Specifically, the value of $m$ reaches its peak within the range of $[3, 5]$. As $m$ increases from 1 to 3, there is a sudden improvement in the model's performance, indicating that the decoupling of relation latent space into multiple perspectives contributes significantly to the construction of relation representations. However, when $m \geq 5$, the model's performance exhibits a gradual decline. This trend suggests that with a higher number of consecutive ``[MASK]" tokens, the prompt-based instance representation obtained by the model tends to contain more noise, thereby adversely affecting the overall model performance.

\begin{figure}[!t]
	\centering
	\includegraphics[width=0.99\linewidth]{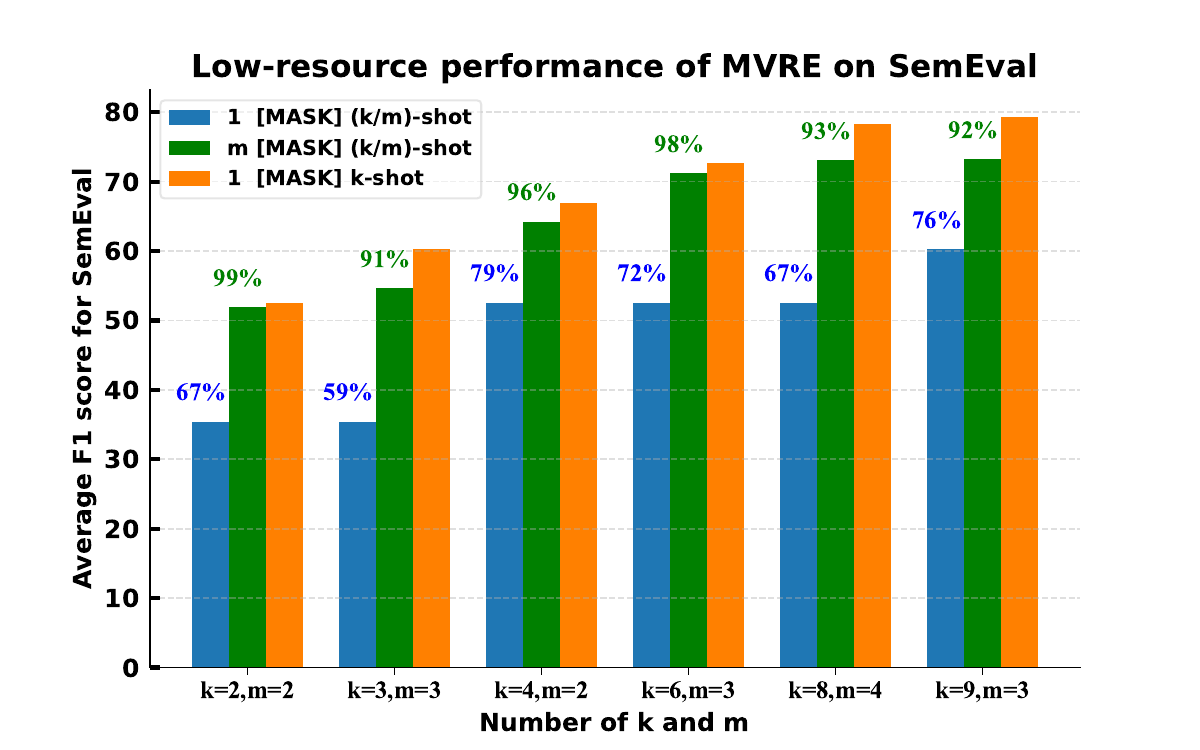}
	\caption{MVRE under low-resource conditions vs. MVRE with only one [MASK] under more resource-rich conditions.}
	\label{fig:few_to_all}
\end{figure}

\subsection{Case study of Dynamic Initialization}

We illustrate the feasibility of multiple ``[MASK]" tokens and the effectiveness of our Dynamic Initialization through a case study, presented in Table~\ref{tab:my-table-init}.

Specifically, for a sentence $x$, we wrap it into $T(x)$ and input $T(x)$ into the model (RoBERTa-large). At each ``[MASK]" position, we obtain the token with the highest probability from the model. This token represents the word that the model identifies as best representing the relation based on the given sentence. During the Dynamic Initialization process, we utilize the embedding of the token with the highest probability to initialize the corresponding position of the virtual relation word.

Given the existence of many relations with reversed subject and object roles in the dataset, it is challenging to model them accurately without confusion. Therefore, in Table~\ref{tab:my-table-init}, we illustrate our method's unique treatment of relations that are mutually passive and active by interchanging the subject and object orders(we controlled the active and passive voice of relations by swapping the order of [sub] and [obj]).
It can be observed that, by increasing the number of [MASK] tokens, RoBERTa-large in the zero-shot scenario effectively captures both active (``was founded in" and ``reflected on") and passive (``the founding of" and ``been reflected in") voice forms for these two relations. However, when there is only one [MASK] token, the generated tokens are largely unrelated to these relations. This indicates that increasing the number of [MASK] tokens enables the PLM to utilize a broader range of words to depict a complex relation, potentially enhancing the PLM's capacity to represent relations.

\subsection{Effectiveness of Low-resource Decoupling Learning}
We conduct experiments to demonstrate the effectiveness of decoupling learning in MVRE, which can be formalized as the following equation in our methods:
$E(\hat H^{\text{1-shot}}_{y_e}) \sim  E(\hat H^{\text{standard}}_{y_e})$. To evaluate the effectiveness of our proposed method, we compare the performance in scenarios with relatively low and enough resources. To be specific, we compare MVRE with one [MASK] against MVRE with $m$ [MASK]. One-[MASK] MVRE is tested in k-shot settings, while m-[MASK] MVRE is tested in (k/m)-shot settings, ensuring consistent relation representation sampling. Additionally, we test one-[MASK] MVRE in (k/m)-shot scenarios for result comparison. The results are as shown in Figure~\ref{fig:few_to_all}. We employ the proportion of model result similarity to represent the overall similarity of obtained relation representations, as represented by the formula:$sim(H\text{-model1}, H\text{-model2})=\frac{\text{F1-score-model1}}{\text{F1-score-model2}}$. Experimental results show that, with an equal number of $h$, the similarity of relation representations obtained under low-resource scenarios surpasses 90\% when compared to higher-resource scenarios. This indicates a 20\% improvement over the one-[MASK] approach. This demonstrates that decoupling relation representations into multi-view perspectives can significantly enhance relation representation capabilities in low-resource scenarios.


\section{Conclusion}
In this paper, we present MVRE for relation extraction, which improves low-resource prompt-based relation representations with multi-view decoupling. Meanwhile, we propose the Global-Local Loss and Dynamic Initialization techniques to constrain the semantics of virtual relation words, optimizing the learning process of relation representations. Experimental results demonstrate that our method significantly outperforms existing state-of-the-art prompt-tuning approaches in low-resource settings.

\section{Acknowledgments}
This work was supported in part by the National Natural Science Foundation of China under Grant No. 62276110, No. 62172039 and in part by the fund of Joint Laboratory of HUST and Pingan Property \& Casualty Research (HPL). The authors would also like to thank the anonymous reviewers for their comments on improving the quality of this paper.



\bibliography{aaai24}

\clearpage
\appendix

\section{A. Hyper-parameters and Reimplemention}
This section details the training and inference process of our models. We train and inference MVRE with PyTorch and Huggingface Transformers on one NVIDIA 4090. All optimizations are performed with the AdamW optimizer. The random seed for data sampling is set to 1 through 5. Due to the utilization of the Dynamic Initialization method, which enhances the initial representation of virtual words' embeddings, we adopt distinct parameters for $\alpha$ and $\beta$ in the Global-Local Loss when using Dynamic Initialization.
\subsection{A.1 Standard Setting}
The hyperparameters of MVRE in the standard-setting experiments are as follows:
\begin{itemize}
    \item learning rate: $5e-6$
    \item batch-size: $8$
    \item max seq length: $256$ (for TACRED, TACREV as $512$)
    \item gradient accumulation steps: 1
    \item number of epochs: 16
    \item $\alpha$: 2 (for using Dynamic Initialization as 1.2)
    \item $\beta$: 0.1 (for using Dynamic Initialization as 0.7)
\end{itemize}
\subsection{A.2 Low-Resource Setting}
The hyperparameters of MVRE in the low-resource setting experiments are as follows:
\begin{itemize}
    \item learning rate: $3e-5$
    \item batch-size: $8$
    \item max seq length: $256$ (for TACRED, TACREV as $512$)
    \item gradient accumulation steps: 1
    \item number of epochs: 40
    \item $\alpha$: 2 (for using Dynamic Initialization as 1.2)
    \item $\beta$: 0.1 (for using Dynamic Initialization as 0.7)
\end{itemize}

\section{B. Visualization of Multi-view Capture}
\begin{figure}[!ht]
    \vspace{-0.4cm}  
    \setlength{\abovecaptionskip}{0.05cm}
	\centering
    \captionsetup{font={scriptsize,stretch=0.7}} %
	\includegraphics[width=1.0\linewidth]{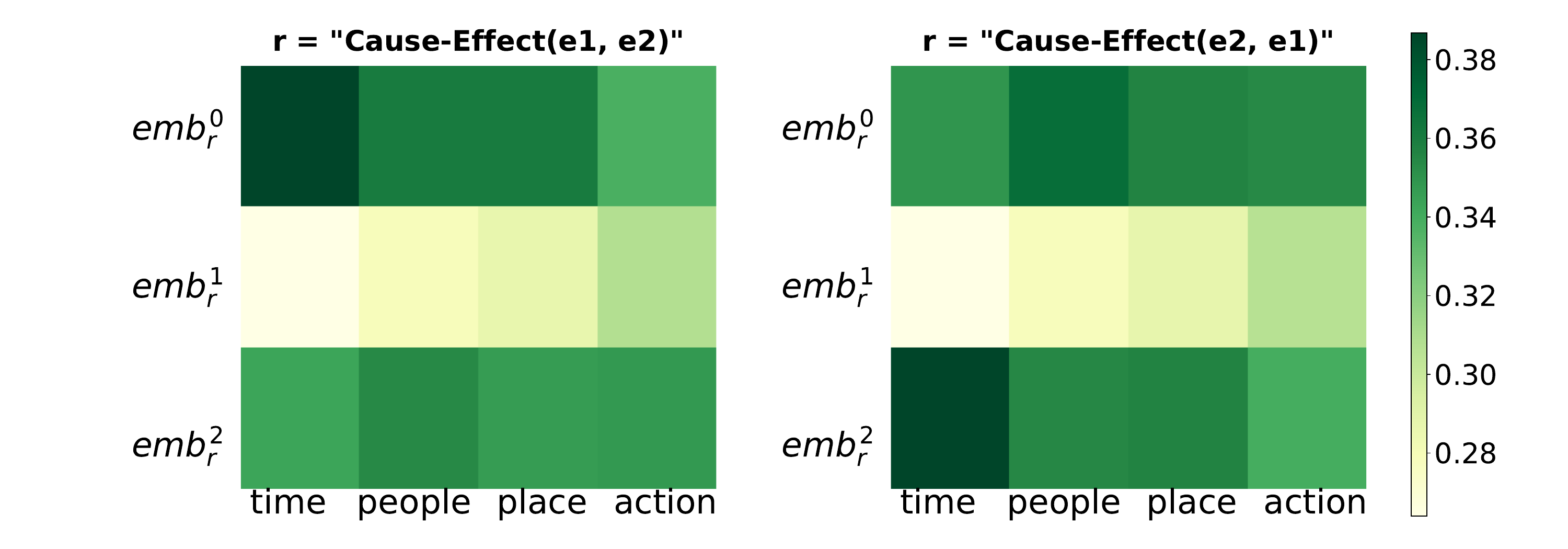}
	\caption{A heat map between different virtual words and aspects. Each row shows how virtual words relate to different views.}
	\label{fig:many_decoupling}
    \vspace{-0.4cm} 
\end{figure}

MVRE is capable of decoupling each complex relation into multiple virtual words, each (i.e., a view) of which denotes a probability distribution over multiple aspects of a complex relation. Then, MVRE jointly optimizes the representations of such multi-views for maximizing the likelihood during inference. In Table~\ref{tab:my-table-init}, only the top-1 result is displayed, with function words selected for their broad semantic coverage. To provide a clearer illustration of the concept of multi-view decoupling, we have designed a special experiment to explore the correlation between different virtual words and various views, such as ``time", ``people", ``place", and ``action". We present a visualization as Figure~\ref{fig:many_decoupling}. In detail, we compute the cosine similarity between each virtual word in MVRE and all non-special words\footnote{Special words: such as [CLS] and other special tokens in the vocabulary, including virtual words} in other vocabulary lists. Then, we calculate the cosine similarity with the top 10 most similar words and words related to ``time", ``people", ``place", and ``action". For example, regarding "time," we can compute the similarity with words such as ``time," ``when," and other temporal descriptors. Finally, the product of these two similarities serves as a measure of relevance between the virtual word and the four specified perspectives: time, people, place, and action. 

\section{C. Manual-constructed Templates for Dynamic Initialization}
In this subsection, we present the manually constructed templates used for Dynamic Initialization in SemEval(Table~\ref{tab:template-semeval}) and TACRED (also applicable to TACREV). During Dynamic Initialization, we utilize RoBERTa-large to predict the word with the highest probability for each [MASK], and then use the embedding of this word to initialize the virtual words corresponding to the respective relations. In the table, $m$ indicates the number of m [MASK].

\begin{table*}[]
\centering
\begin{tabular}{cc}
\hline
{Relation}            & Template                                                               \\ \hline
Member-Collection(e1,e2)  & member is in the collection. member {[}MASK{]}*m collection.           \\
Member-Collection(e2,e1)  & collection is a set of members. collection {[}MASK{]}*m members.       \\
Product-Producer(e1,e2)   & product is made by producer. product {[}MASK{]}*m producer.            \\
Product-Producer(e2,e1)   & producer make out a product. producer {[}MASK{]}*m product.            \\
Entity-Origin(e1,e2)      & entity derived from the origin. entity {[}MASK{]}*m origin.            \\
Entity-Origin(e2,e1)      & origin is the source of entity. origin {[}MASK{]}*m entity.            \\
Cause-Effect(e1,e2)       & cause that causes to effect. cause {[}MASK{]}*m effect.                \\
Cause-Effect(e2,e1)       & effect is caused by cause. effect {[}MASK{]}*m cause.                  \\
Entity-Destination(e1,e2) & the target of entity is destination . entity {[}MASK{]}*m destination. \\
Entity-Destination(e2,e1) & destination is the target of entity. destination {[}MASK{]}*m entity.  \\
Component-Whole(e1,e2)    & component is in the whole. component {[}MASK{]}*m whole.               \\
Component-Whole(e2,e1)    & whole is comprised of components. whole {[}MASK{]}*m components.       \\
Content-Container(e1,e2)  & content is in container. content {[}MASK{]}*m container.               \\
Content-Container(e2,e1)  & container is containing the content. container {[}MASK{]}*m content.   \\
Message-Topic(e1,e2)      & message is about the topic. message {[}MASK{]}*m topic.                \\
Message-Topic(e2,e1)      & topic is described through message. topic {[}MASK{]}*m message.        \\
Instrument-Agency(e1,e2)  & instrument is used by agency. instrument {[}MASK{]}*m agency.          \\
Instrument-Agency(e2,e1)  & agency using the instrument. agency {[}MASK{]}*m instrument.           \\
Other                     & subject and object are not related. subject {[}MASK{]}*m object.                             \\ \hline
\end{tabular}
\caption{The template used for Dynamic Initialization in SemEval.}
\label{tab:template-semeval}
\end{table*}

\begin{table*}[]
\centering
\resizebox{\linewidth}{!}{
\begin{tabular}{cc}
\hline
{Relation}            & Template                                                               \\ \hline
per:title & subject person title object. subject {[}MASK{]}*m object. \\
per:employee\_of & subject person employee of object. subject {[}MASK{]}*m object. \\
NA & subject no relation object. subject {[}MASK{]}*m object. \\
per:countries\_of\_residence & subject person countries of residence object. subject {[}MASK{]}*m object. \\
org:top\_members/employees & subject organization top members or employees object. subject {[}MASK{]}*m object. \\
org:country\_of\_headquarters & subject organization country of headquarters object. subject {[}MASK{]}*m object. \\
per:religion & subject person religion object. subject {[}MASK{]}*m object. \\
per:cause\_of\_death & subject person cause of death object. subject {[}MASK{]}*m object. \\
org:alternate\_names & subject person alternate names object. subject {[}MASK{]}*m object. \\
per:city\_of\_birth & subject person city of birth object. subject {[}MASK{]}*m object. \\
per:cities\_of\_residence & subject person cities of residence object. subject {[}MASK{]}*m object. \\
org:city\_of\_headquarters & subject organization city of headquarters object. subject {[}MASK{]}*m object. \\
per:age & subject person age object. subject {[}MASK{]}*m object. \\
per:city\_of\_death & subject person city of death object. subject {[}MASK{]}*m object. \\
per:origin &subject person origin object. subject {[}MASK{]}*m object. \\
per:other\_family & subject person other family object. subject {[}MASK{]}*m object. \\
org:subsidiaries & subject organization subsidiaries object. subject {[}MASK{]}*m object. \\
per:children & subject person children object. subject {[}MASK{]}*m object. \\
org:dissolved & subject organization dissolved object. subject {[}MASK{]}*m object. \\
per:stateorprovinces\_of\_residence & subject person state or provinces of residence object. subject {[}MASK{]}*m object. \\
per:siblings & subject person siblings object. subject {[}MASK{]}*m object. \\
per:spouse & subject person spouse object. subject {[}MASK{]}*m object. \\
per:stateorprovince\_of\_death & subject person state or province of death object. subject {[}MASK{]}*m object. \\
per:alternate\_names & subject person alternate names object. subject {[}MASK{]}*m object. \\
org:member\_of & subject organization member of object. subject {[}MASK{]}*m object. \\
org:parents & subject organization parents object. subject {[}MASK{]}*m object. \\
org:website & subject organization website object. subject {[}MASK{]}*m object. \\
per:parents & subject person parents object. subject {[}MASK{]}*m object. \\
org:founded & subject organization founded object. subject {[}MASK{]}*mobjectB. \\
org:stateorprovince\_of\_headquarters & subject organization state or province of headquarters object. subject {[}MASK{]}*m object. \\
per:schools\_attended & subject person schools attended object. subject {[}MASK{]}*m object. \\
org:members & subject organization members object. subject {[}MASK{]}*m object. \\
org:political/religious\_affiliation & subject organization political or religious affiliation object. subject {[}MASK{]}*m object. \\
per:date\_of\_birth & subject person date of birth object. subject {[}MASK{]}*m object. \\
org:founded\_by & subject organization founded by object. subject {[}MASK{]}*m object. \\
org:shareholders & subject organization shareholders object. subject {[}MASK{]}*m object. \\
org:number\_of\_employees/members & subject organization number of employees or members object. subject {[}MASK{]}*m object. \\
per:country\_of\_birth & subject person country of birth object. subject {[}MASK{]}*m object. \\
per:stateorprovince\_of\_birth & subject person state or province of birth object. subject {[}MASK{]}*m object. \\
per:charges & subject person charges object. subject {[}MASK{]}*m object. \\
per:date\_of\_death & subject person date of death object. subject {[}MASK{]}*m object. \\
per:country\_of\_death & subject person country of death object. subject {[}MASK{]}*m object. \\ \hline
\end{tabular}
}
\caption{The template used for Dynamic Initialization in TACRED (also utilized in TACREV).}
\label{tab:template-tacred}
\end{table*}

\end{document}